\documentclass[9pt,twocolumn,twoside]{pnas-report}
\templatetype{pnasresearcharticle}

\usepackage{lipsum}

\title{Extractive approach for text summarization using graphs}

\author[a,1]{Kastriot Kadriu}
\author[a]{Milenko Obradovic} 

\affil[a]{University of Ljubljana, Faculty of Computer and Information Science, Ve\v{c}na pot 113, SI-1000 Ljubljana, Slovenia}
\leadauthor{Kadriu} 

\authordeclaration{All authors contributed equally to this work.}
\correspondingauthor{\textsuperscript{1}To whom correspondence should be addressed. E-mail: kk5222@student.uni-lj.si.}
\begin{abstract}
Natural language processing is an important discipline with the aim of understanding text by its digital representation, that due to the diverse way we write and speak, is often not accurate enough. Our paper explores different graph related algorithms that can be used in solving the text summarization problem using an extractive approach. We consider two metrics: sentence overlap and edit distance for measuring sentence similiarity. Relevant structures have been implemented and the code can be obtained in this link \cite{TOTI}.
\end{abstract}

\dates{The manuscript was compiled on \today}
\doi{\href{https://ucilnica.fri.uni-lj.si/course/view.php?id=183}{Introduction to Network Analysis} 2020/21}

\begin{document}

\maketitle
\begin{figure}[t]\centering%
	\includegraphics[width=1\linewidth]{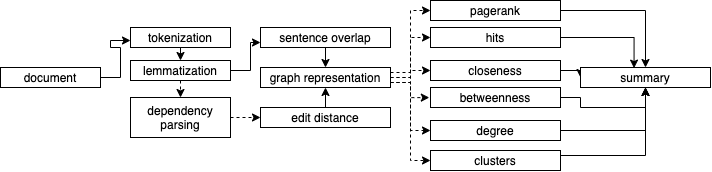}
	\caption{Pipeline of our model}
	\label{fig:pipeline}
\end{figure}

\thispagestyle{firststyle}
\ifthenelse{\boolean{shortarticle}}{\ifthenelse{\boolean{singlecolumn}}{\abscontentformatted}{\abscontent}}{}
\nocite{*}
\dropcap{W}hile the quantity of information is growing exponentially, there's a need to compress the content in condensed versions. The summarization of information is a problem that deals with presenting the main idea of the text without the need to read all the content of the text. Such task is heavily relied on the calculation of sentence similiarity, and in that regard, various methods have been tried. \\
Two methods for automatic text summarization are considered: extractive and abstractive. Extractive summarization is based on the identification of important sections of the text and producing a subset of the sentences from the original text, whereas abstractive summarization tries to reproduce the important content in a new way after interpretation and examination of text using more advenced techniques. \\
Supervised summarization models are built by treating the problem as a classification task, and by classifying which features in a sentence are relevant for summarization. \\ Those models aren't very reliable because of the unpredictable nature of language, from which it is not easy to generate a classification pattern. In addition, such models require training data which worsen the data acquisition problem which is already a challenge on its own. \\ 
Task summarization problem concerns itself with representing data in such way that the importance of each sentence and their terms is properly considered. Text should be represented in such ways that the inter-word and inter-sentence dependency is kept.\\
The task is challenged by a sustainable data source for validation and subsequently, an efficient metric for evaluating it and other text-understanding tasks. The challenges are topics of Document Understanding Conference, which later became Text Analaysis Conference. A domain-independent evaluation is also complex to be achieved. For example, a model reporting a high accuracy in summarizing news articles might not perform with the same accuracy when summarizing, let's say, Reddit posts. \\ 
Generally, best performing models are deep learning based. The construction of such models is faced with challenges of its own mainly regarded with computational resources. For that reason, it's important to consider more simpler approaches like the ones that are presented in this paper. \\
Our contribution includes a direct implementation of graph-based algorithms for computing relevant data that could summarize the test documents the best. The respective algorithms help us extract the most important sentences that will constitute the summary. We test two different metrics for computing sentence similiarity, and in one of them we employ a notion of graph similiarity - edit distance - Figure \ref{fig:pipeline}.

\section*{Related work}
Text summarization is an open problem. There hasn't been report of an official model who can achieve a decent data-independent accuracy. Current state-of-art models achieve accuracy of around 50\% percent. Those models are usually deep learning models. \cite{YW} achieves state-of-the-art results on the CNN/Daily mail dataset. The model presented there is the Reinforced Neural Extractive Summarization (RNES) model. \cite{TSK} presents a general overview of two ranking algorithms - PageRank and HITS, and an agnostic overview of building a graph representation of text to be used for summarization by extraction. In addition, for smooth-er outputs, shortest path algorithm is suggested. This paper lacks concrete results tested on some dataset. \\
\cite{MH, MCH} dive deeper in the use of PageRank algorithm for text summarization, whose use in such cases is referred to as TextRank. 
Sentences are extracted using the respective algorithm in a weighted graph built with nodes representing the sentences to be summarized, and weighted edges represent the similarity those sentences have with each other. Similiarity between sentences is calculated as their overlap which can be determined as the number of common words between their lexical representation. The resulting model is an unsupervised model that has achieved 47\% accuracy, as evaluted by the ROGUE metric, on 567 news articles provided by the Document Understanding Evaluations (DUC) 2002. The paper also explores the use of TextRank in keyword extraction. The paper introduces the respective algorithm very well but could benefit by the computation of a larger batch of test data. \\

\cite{KM} implements the TextRank algorithm and expands on the pre and post-processing part. The data (sentences) is encoded and different methods are considered such as tf-idf and Word2vec. Moreover, the implementation is tested on Malayaian content, demonstrating the domain independence of the algorithm. 
Another extractive text summarization algorithm is LexRank, which is based on computing the importance of a sentence by the concept of eigenvector centrality presented in \cite{EG}. The similiarity between sentences, also represented as nodes, is calculated as the cosine similiarity between the vector of their words mapped as their \textit{idf} values. \\
Topic based approaches are seen on \cite{KARI, NG, RH}. They are based on the distribution of words accross documents from which it is possible to derive \textit{topics} that later constitute summaries.
\cite{KATJA} treats the problem of summarization as a compression problem that could be integrated in both extractive and abstractive summarization approaches, and \cite{MM} builds a small world network to summarize biomedical articles.

\section*{Results}
Native structures have been implemented with the aim of supporting experiments. Graphs have been created using NetworkX \cite{nx}, whereas linguistic processing has been done with nltk \cite{NLTK} and stanza \cite{qi2020stanza}. \\
The aim of our experiments was to test how well different centrality measures are able to identify the most important parts of the text. For that reason, we consider pagerank, hits, closeness, betweenness and degree measures. In addition, when construction the edges of the graph, we use two different metrics for measuring how similar two sentences are, thus determing the weight of the link between those two nodes: word overlap and edit distance. The edge candidates are then ranked based on their weight, and then, by a threshold value, the edges with the top weights are created. \\
The model has been evaluated using \textbf{3500} documents from the CNN/Dailymail \cite{DBLP:journals/corr/SeeLM17, hermann2015teaching} dataset - Table \ref{tbl:dataset}, which is a collection of news articles with their highlights serving as our summarization tests. 
\begin{table}[t]\centering%
	\caption{CNN/Dailymail dataset statistics.}
	\begin{tabular}{lcc}\toprule
	    Number of test documents & 3500\\
		\midrule
		Average document length & 33 sentences \\
		\midrule
		Average summary length & 5 sentences \\
		\midrule
		Average document-summary compression & 0.85\\ \bottomrule
	\end{tabular}
	\label{tbl:dataset}
\end{table}

Various methods have been tested with two different metrics: sentences overlap and sentences edit distance (TED). After employing strategies described in the Methods section, top five sentences, number which is based on the average summary length, are extracted that represent the generated summary with our method(s). The summary is then compared with the ground truth summary provided in the dataset using the ROGUE metric. \\
Edit distance for sentence similiarity is more time consuming because it adds three extra steps in our pipeline: word dependency parsing, tree construction and edit distance calculation. \\
We report on recall and F1 score values. We are using recall to evaluate how much of the 'correct' content is included in summaries, and F1 score to give a balanced evaluation score that is not based on the attributes that recall takes into account such as the summary length. 
\begin{table}[t]\centering%
	\caption{Recall values of the evaluation using the sentence overlap metric and t=0.5.}
	\begin{tabular}{lccc}\toprule
	   	\textbf{Method} & \textbf{Rogue-1} & \textbf{Rogue-2} & \textbf{Rogue-L}\\
		\midrule
		Pagerank & 0.50 & 0.19 & 0.44 \\
		\midrule
		Hits & 0.48 & 0.19 & 0.42 \\
		\midrule
		Closeness & 0.50 & 0.20 & 0.45 \\ 
		\midrule
		Betweenness & 0.50 & 0.20 & 0.44 \\
		\midrule
		Degree & 0.50 & 20 & 0.45 \\
		\midrule
		Clusters & 0.47 & 0.18 & 0.42 \\
		\bottomrule

	\end{tabular}
	\label{tbl:metric_1}
\end{table}

\begin{table}[t]\centering%
	\caption{Recall values of the evaluation using TED metric and t=0.5.}
	\begin{tabular}{lccc}\toprule
	   	\textbf{Method} & \textbf{Rogue-1} & \textbf{Rogue-2} & \textbf{Rogue-L}\\
		\midrule
		Pagerank & 0.48 & 0.17 & 0.426 \\
		\midrule
		Hits & 0.48 & 0.17 & 0.423 \\
		\midrule
		Closeness & 0.51 & 0.19 & 0.45 \\ 
		\midrule
		Betweenness & 0.51 & 0.19 & 0.45 \\
		\midrule
		Degree & 0.51 & 0.19 & 0.45 \\
		\midrule
		Clusters & 0.43 & 0.14 & 0.38 \\
		\bottomrule

	\end{tabular}
	\label{tbl:metric_2}
\end{table}

\begin{table}[t]\centering%
	\caption{F score values of the evaluation using the sentence overlap metric and t=0.5.}
	\begin{tabular}{lccc}\toprule
	   	\textbf{Method} & \textbf{Rogue-1} & \textbf{Rogue-2} & \textbf{Rogue-L}\\
		\midrule
		Pagerank & 0.26 & 0.10 & 0.28 \\
		\midrule
		Hits & 0.26 & 0.10 & 0.28 \\
		\midrule
		Closeness & 0.28 & 0.12 & 0.30 \\ 
		\midrule
		Betweenness & 0.28 & 0.11 & 0.29 \\
		\midrule
		Degree & 0.28 & 0.11 & 0.29 \\
		\midrule
		Clusters & 0.27 & 0.10 & 0.27 \\
		\bottomrule

	\end{tabular}
	\label{tbl:metric_1_fscore}
\end{table}

\begin{table}[t]\centering%
	\caption{F score values of the evaluation using TED metric and t=0.5.}
	\begin{tabular}{lccc}\toprule
	   	\textbf{Method} & \textbf{Rogue-1} & \textbf{Rogue-2} & \textbf{Rogue-L}\\
		\midrule
		Pagerank & 0.24 & 0.08 & 0.244 \\
		\midrule
		Hits & 0.24 & 0.08 & 0.246 \\
		\midrule
		Closeness & 0.24 & 0.09 & 0.25 \\ 
		\midrule
		Betweenness & 0.24 & 0.09 & 0.25 \\
		\midrule
		Degree & 0.24 & 0.09 & 0.25 \\
		\midrule
		Clusters & 0.25 & 0.08 & 0.25 \\
		\bottomrule

	\end{tabular}
	\label{tbl:metric_2_fscore}
\end{table}
The results from Table \ref{tbl:metric_1}, \ref{tbl:metric_2}, \ref{tbl:metric_1_fscore} and \ref{tbl:metric_2_fscore} might not have very seductive numbers but they are standard numbers across summarization models and are actually comparable with the state-of-the-art models, like the one in \cite{YW}, although we can not make direct comparisons because their model has been tested in a greater batch of testing data than ours. 
Rogue-1 tells us the number of golden tokens found in the summary. Rogue-2 tells us the number of bigrams that were matched between test and ground truth summary but the values here aren't expected to be high as the ordering of the sentences (and words, although since our approach is extractive, that's not a big issue for our case) matters. Rouge-L is an important value to consider as it measures the relation of the two contents in a wider context.

\section*{Discussion}
The performance of closeness, betweenness and degree centrality measures is approximate with each other. It is completely affected by the threshold value that determines the number of edges to be created. Hypothetically, if the threshold is 1, the output of the summary would be the first 5 lines of the article. In addition, using those methods, there is a limitation on the possibilities for fine tuning the results. \\
Pagerank and hits perform the most consistently, and that is based on the fact that those algorithms take into account the weight of the edges. Those methods aren't affected by the number of generated edges, as long as the weights are properly accounted for. Those methods are in the same spirit, they both formalise link analysis as eigenvector problems. Pagerank reports higher accuracy because it is able to operate in the complete graph, unlike Hits which operates on a smaller subgraph. In addition, those methods work well because they do not rely on the isolated information regarding the node but instead, they take into account the entire graph and the relationships within. \\
The clustering method offers more room for tuning the results. The clustering process is based on finding cliques which is a hard problem. It is more computationally expensive compared to other algorithms. It performs best with the tree edit distance as this measure is able to represent the similiarity of sentences for clustering purposes. It is affected by the threshold parameter. \\
The lack of higher numbers in the results is not neccessarily because of the actual quality of our summaries, rather than the limitation that ROGUE metric has. ROGUE address content selection between the test and the grouth truth content without accounting for other quality aspects such as coherence, gramamaticality or fluency. The content selection is relied on lexical overlap but a good summary isn't always expressed with the same lexical links, and this happens especially in abstractive summarization. It would have been nice if there were more ground truth summaries to compare our generated summaries with, as they could constitute a more leveled evaluation. The dataset we used, however, provided only one summary per article. \\ 
Our paper offers an overview in using graphs to represent the relation that sentences have, relevant algorithms for detecting the relationships within the graph and presenting them but it could be improved in several ways.
First, a more natural way of picking the top relevant sentences for summary can be integrated, instead of arbitrary picking top $N$ sentences.
Then, to have results tailored for specific purposes, the sentence similiarity metrics could have been modified to report higher scores for the presence of certain structures in the sentences, such as noun, verbs or adjectives.  \\ 
The clustering method could be favorable as we can incorporate a hybrid approach towards summarization by turning the problem into a graph compression problem after finding the relevant clusters, where for each cluster we try to compress it into a single sentence by linking the most important part of sentences. 
Other than text summarization, the findings in this paper can also be referenced as proposals for solving classification problems using graphs. \\
The work on text summarization problem needs to be supported by a standard dataset that covers a more diverse textual content. Future work should also approach new metrics for content evaluation that could overcome ROGUE limitations. Such metrics should account for the relation that words have with each other, preferrably through an embedding methodology and based on an external corpus, although in the first sight this already looks computationally heavy and that's another issue that would need to be taken care off.

{\small\section*{Methods}
\subsection*{Pre-processing}
Sentences of a document have been tokenized, cleaned up and the next processes have been executed depending on the metric. For the sentence similiarity, the words in sentences are represented by their lemmas so that the context is not missed even if words are in different forms. For the edit distance, we build dependency parsed trees for each sentence using the stanford parser (stanza).
\subsection*{Metrics}
One metric used for measure sentence similiarity is by checking their overlapping words \cite{MH}. To avoid bias on long sentences, the number of overlapping words is normalised by the lengths of both sentences. 
\[
Sim(S_{i}, S_{j}) = \frac{|\{w_{k}|w_{k} \in S_{i} \& w_{k} \in S_{j}\}|}{log(|S_{i}| + log(|S_{j}|))}	
\]
The other metric used to measure how similar two sentences are is the edit distance, in our case since the sentences are represented as tree, we work with tree edit distance. The tree edit distance (TED) \cite{GSGA} is calculated using the algorithm proposed by Zhang and Shasha in \cite{ZSSS}. TED is a more accurate metric because it is able to account for the number of words in a sentence needed to be changed to match the other sentence, as compared to string based distance metrics that report on character changes, and such metric would be totally incorrect when comparing sentences of different lengths. The similarity of two sentences based on their distance is calculated as: 
\[
Sim(i, j) = \frac{1}{1 + d(T_{i}, T_{j})}	
\]
\subsection*{Graph construction}
After the document has been processed into sentences, we construct the undirected graph by generating a node for each sentence. Then, similiarities between each pair of sentences is calculated. Afterwards, the similiarities are sorted from highest to lowest, and based on a user defined threshold $t$, we pick the top $t$ percent of similiar pairs to serve as the edges of the graph. 

\begin{figure}[b]\centering%
	\includegraphics[width=0.40\linewidth]{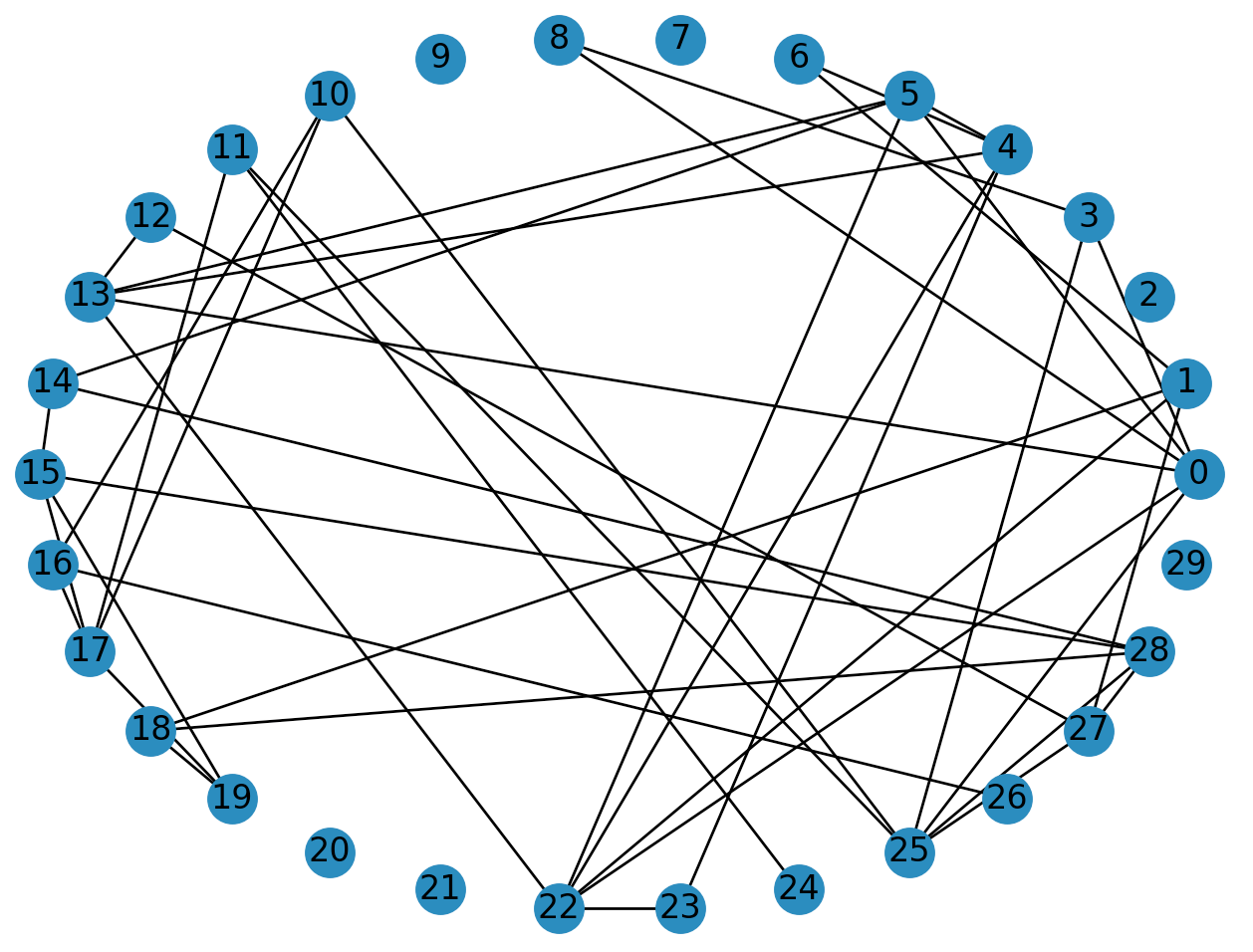}\hskip12pt
	\includegraphics[width=0.40\linewidth]{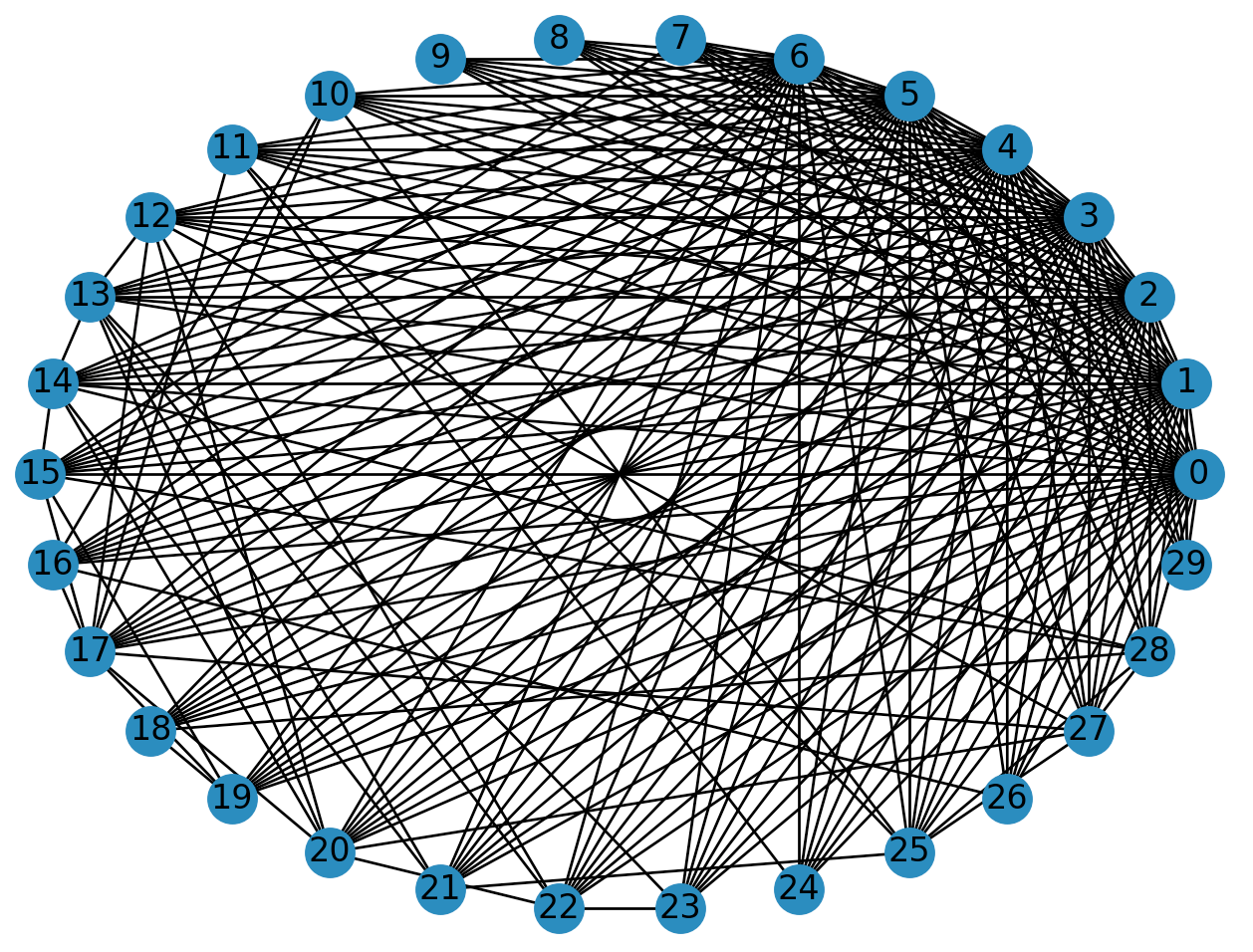}
	\caption{Sentence similiarity graphs generated using t=0.1 and t=0.5 respectively.}
	\label{fig:example}
\end{figure}
\subsection*{Algorithms}
The implementation of PageRank, Hits, Closeness, Betweenness and degree measures is straight forward - after the scores for each node(sentence) have been computed, they are ranked and top N sentences are picked where $N$ is a user defined parameter symbolizing the desired summary length. Whereas, for the clustering method, after the cliques in the graph have been found we process them in the following way: \begin{itemize}
	\item we ignore cliques with one item
	\item for cliques with more than two items, we consider their closeness scores from which we pick the one item with the highest scores
	\item in case one item is found to be the highest scoring item in more than one clique, we pick that as the representive for the cluster with the bigger clique, and the selection moves onto the second highest scoring item in the remaining clique
\end{itemize} 
\subsection*{Evaluation}
Our extracted summaries are evaluated using the Recall-Oriented Understudy for Gisting Evaluation - ROGUE metric. We have incorporated three types of  ROGUE evaluation metric. Rogue-1, the overlap of unigrams (each word) between extracted summary and ground truth summary provided in the dataset, Rogue-2, the overlap of bigrams in the respective relation, and Rogue-L which reports on the longest common subsequence statistics.
\acknow{The authors would like to acknowledge these works:}

\showacknow{}
\bibliography{bibliography}

\end{document}